\documentclass[review,12pt]{elsarticle}

\usepackage{amssymb,graphicx}
\usepackage{amsmath}
\usepackage{booktabs}
\usepackage{multirow}
\usepackage{warpcol}
\usepackage{subfigure}
\usepackage{rotating}
\usepackage{pdflscape}
\usepackage{natbib}
\usepackage{lineno}
\usepackage{flushend}
\usepackage{float}
\usepackage{latexsym}
\usepackage{caption}
\usepackage{longtable}
\usepackage{setspace}
\usepackage{geometry}
\usepackage{color}
\usepackage{threeparttable}
\usepackage{algorithm}
\usepackage{algorithmic}

\journal{Science of Remote Sensing}
\begin{document}
\begin{frontmatter}



\title{A Multi-Source Data Fusion-based Semantic Segmentation Model for Relic Landslide Detection} 


\author[1]{Yiming Zhou}
\ead{zhouyiming@bupt.edu.cn}

\author[1]{Yuexing Peng\corref{cor1}}
\ead{yxpeng@bupt.edu.cn}

\author[2]{Daqing Ge}

\author[2]{Junchuan Yu}

\author[3]{Wei Xiang}

\cortext[cor1]{Corresponding author.}
\affiliation[1]{organization={Key Lab of Universal Wireless Communication, MOE, Beijing University of Posts and Telecommunications},
            city={Beijing},
            postcode={100876}, 
            country={China}}

\affiliation[2]{organization={China Aero Geophysical Survey and Remote Sensing Center for Natural Resources},
            city={Beijing},
            postcode={10083},
            country={China}}

\affiliation[3]{organization={School of Engineering and Mathematical Sciences, La Trobe University},
            city={Melbourne},
            postcode={3086},
            state={Victoria},
            country={Australia}}
\begin{abstract}
As a natural disaster, landslide often brings tremendous losses to human lives, so it urgently demands reliable detection of landslide risks. When detecting relic landslides that present important information for landslide risk warning, problems such as visual blur and small-sized dataset cause great challenges when using remote sensing images. To extract accurate semantic features, a hyper-pixel-wise contrastive learning augmented segmentation network (HPCL-Net) is proposed, which augments the local salient feature extraction from boundaries of landslides through HPCL and fuses heterogeneous information in the semantic space from high-resolution remote sensing images and digital elevation model data. For full utilization of precious samples, a global hyper-pixel-wise sample pair queues-based contrastive learning method is developed, which includes the construction of global queues that store hyper-pixel-wise samples and the updating scheme of a momentum encoder, reliably enhancing the extraction ability of semantic features. The proposed HPCL-Net is evaluated on the Loess Plateau relic landslide dataset and experimental results verify that the proposed HPCL-Net greatly outperforms existing models, where the mIoU is increased from 0.620 to 0.651, the Landslide IoU is improved from 0.334 to 0.394 and the F1score is enhanced from 0.501 to 0.565.\end{abstract}

\begin{highlights}
  \item Propose HPCL-Net for relic landslide detection to address the visual blur problem.
  \item Design a dual-branch module to effectively extract and fuse heterogeneous features.
  \item Design hyper-pixel-wise contrastive learning method for small-sized dataset problem.
  \item Build global category queues achieving cross-image contrast to enhance effectiveness.
\end{highlights}


\end{frontmatter}



\section{Introduction}
Landslides are severe geological hazards that often results in traffic disruption, village burial, river blockage and other catastrophic incidents, posing significant threats to human life and property. For instance, Lanzhou City, China, has experienced 24 large-scale landslides since 1949, resulting in 670 fatalities and direct economic losses of 776 million RMB \citep{LandslideLP}. Therefore, accurate and efficient landslide prediction is crucial for preventing and reducing such disastrous losses. Besides, studying existing landslides could provide valuable insights for predicting potential landslides that lack fully-formed features and present the greatest risks, so this paper focuses on improving the detection of existing landslides.

Traditional methods for landslide detection rely on field investigations and remote sensing data, which, though reliable, are time-consuming, inefficient, and highly dependent on experts experience, therefore hard to be widely applied. With the development of remote sensing technologies, abundant ground-observable data has been introduced for automatic landslide detection. For example, Interferometric Synthetic Aperture Radar (InSAR) data can provide deformation characteristics \citep{liang2023automatic, zhou2023time}, High-Resolution Satellite Image (HRSI) data can offer optical features \citep{lu2023iterative, liu2023feature}, and Digital Elevation Model (DEM) and Digital Surface Model (DSM) data can present topographic information \citep{Ruijie2018DEM, DSMFusion2019}. There are two classic methods using HRSI: pixel-based classification and object-oriented analysis (OBA). Pixel-based methods classify pixels based on spectral features \citep{li2016landslide, 7855696, keyport2018comparative} but suffer from salt-and-pepper noise and lack the ability to identify spatially continuous regions \citep{stumpf2011object, prakash2020mapping}, whereas OBA methods improve the spatial continuity by grouping adjacent pixels into segments or objects with a homogeneity factor, leveraging spatial, texture, context, geometry, and spectral features \citep{blaschke2010object, martha2010characterising, lahousse2011landslide, holbling2012semi, pawluszek2019multi}. However, OBA methods require manual adjustments for thresholds such as relief and slopel, limiting their scalability. 

Recent machine learning methods including support vector machine, random forest, logistic regression \citep{chen2018practical, dou2020improved, nhu2020landslide}, and convolutional neural networks (CNNs) have demonstrated improved accuracy and efficiency in landslide detection tasks. CNNs in particular are widely applied due to their excellent end-to-end self-learning ability and abstract representation capability. In \cite{fang2020gan} a Siamese CNN based on dual-temporal landslide HRSI data is developed for the pixel-wise change detection of landslides and a generative adversarial network is introduced to suppress differences between two images in the time domain. In \cite{ji2020landslide}, a 3D spatial and channel attention module is designed to enhance the extraction of landslide features. In \cite{wang2023dual}, a dual path attention network is proposed to leverage both texture and spatial information in remote sensing images.

\begin{figure}[h]
  \centering
  \includegraphics[scale=0.45]{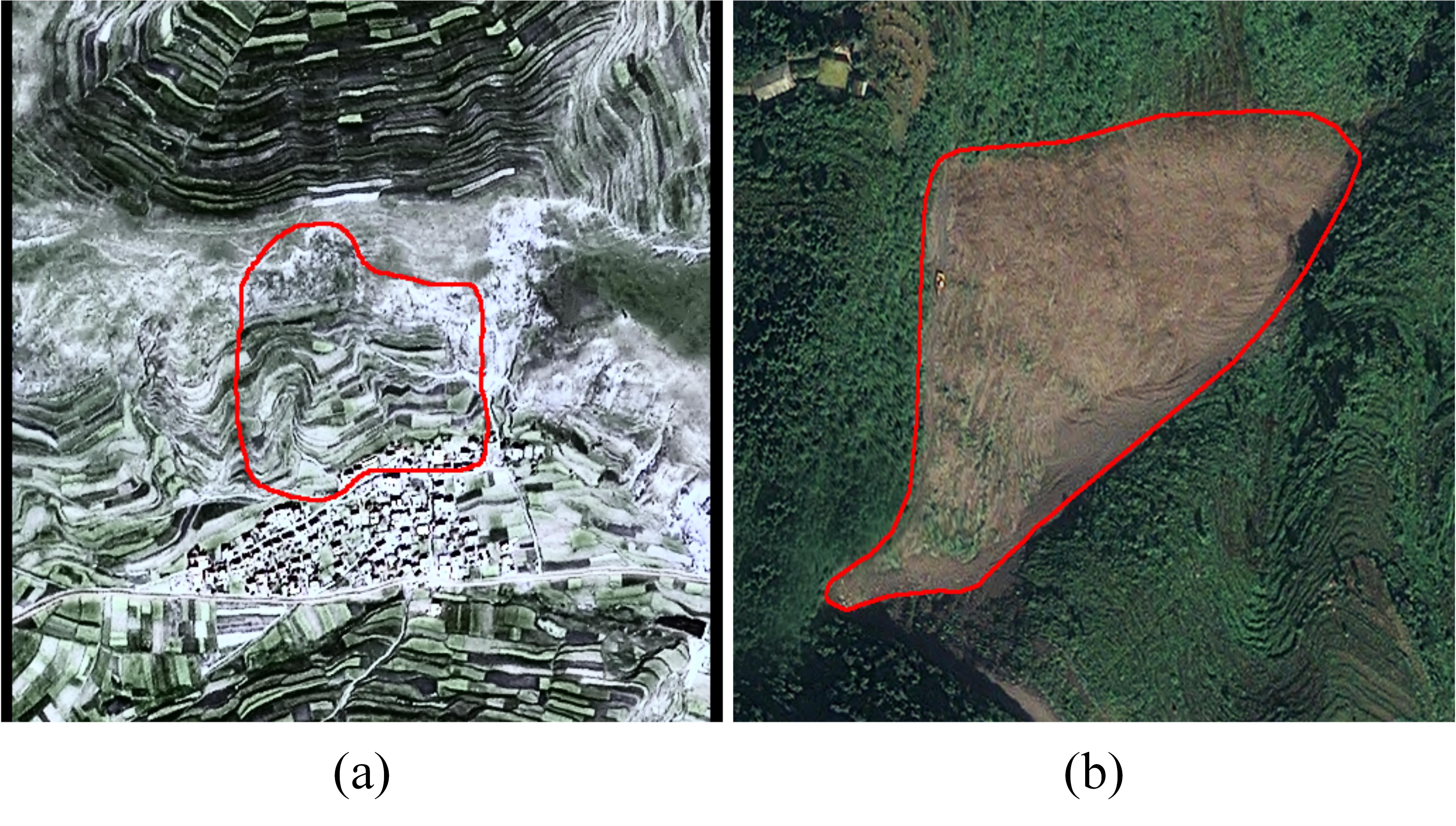}
  \DeclareGraphicsExtensions.
  \caption{An example to show the visual differences between an relic landslide and a new landslide. (a) relic landslide, (b) new landslide.}
  \label{OldandNew}
\end{figure}

Despite significant progress, most research focuses on detecting new landslides \citep{fang2020gan, ghorbanzadeh2019evaluation, ji2020landslide, ju2020automatic, 9428612, wang2023dual} rather than relic landslides \citep{ZHANGYong, du2021landslide, liu2023feature,lu2023iterative}. Since landslide morphology gradually changes over time, there is a significant difference in morphology between new and relic landslides. New landslides, in a state of repeated or just stopped activity, present visually obvious features, such as distinct colors and contours against background. Relic landslides, however, have been inactive for a long time and are severely affected by vegetation coverage, natural erosion, and human activities such as terracing, building houses, and constructing roads, exhibiting blurry visual features difficult to distinguish from background, as shown in Fig. \ref{OldandNew}. Therefore, relic landslides are much difficult to identify than new ones.

To address this challenge, we propose a multi-modal data fusion method leveraging HRSI and DEM data, where DEM data can provide elevation features and gradient patterns that effectively complement HRSI’s optical information. The common data fusion strategies can be categorized into early fusion \citep{ghorbanzadeh2019evaluation, ji2020landslide} and late fusion \citep{DSMFusion2019, MultiModal2019, Zeng2020,  Honghui2023novel, liu2023feature}. The early fusion integrates multi-modal data at preprocessing stage such as combining several inputs from different modalities into one or mixing several datasets as a whole. For example, in \cite{ghorbanzadeh2019evaluation}, topographic images are regarded as additional channels of spectral images, and in \cite{ji2020landslide}, satellite optical images, shape files of landslides' boundaries and DEM data are mixed into one integral dataset. The early fusion can provide richer information and is easy to implement, but it requires powerful models, large training dataset, and massive computational resources, making it unsuitable for scenarios involving small-sized datasets. Besides, due to the fact that different modalities of data often have different semantic and syntactic features (e.g., visual features such as morphology, texture, and color from HRSIs, and terrain features like height variation from DEM), the data space after fusion is always complex and irregular, thus making further feature extraction from it quite challenging. This calls for more powerful models which place higher demands on datasets. In contrast, the late fusion independently extract heterogeneous features and then fuse them in the semantic space, which effectively obtains the essential semantic information required for the task. For example, in \cite{DSMFusion2019}, the semantic features from near infrared, red, green (IRRG) spectrum and Digital Surface Model (DSM) data are concatenated in the channel dimension; in \cite{MultiModal2019}, optical image features and DSM data features are contenated; in \cite{Zeng2020}, two regional feature maps from optical image and Synthetic Aperture Radar data are contenated; in \cite{liu2023feature}, the features from HRSI and DEM data are fused by adding pixels with the same positions in spatial dimension. Therefore, considering the heterogeneity between HRSIs and DEM data, we adopt the late fsuion. Instead of pixel-wise addition \citep{liu2023feature}, we fuse by channel concat since it performs better when there are significant semantic differences between heterogeneous features, such as HRSI and DEM, whereas the pixel-wise addition increases less complexity but at the expense of the discriminative ability of the model for heterogeneous features.

\begin{figure}[h]
    \setlength{\abovecaptionskip}{0.cm}
    \setlength{\belowcaptionskip}{-0.cm}	
    \centering
    \includegraphics[scale=0.4]{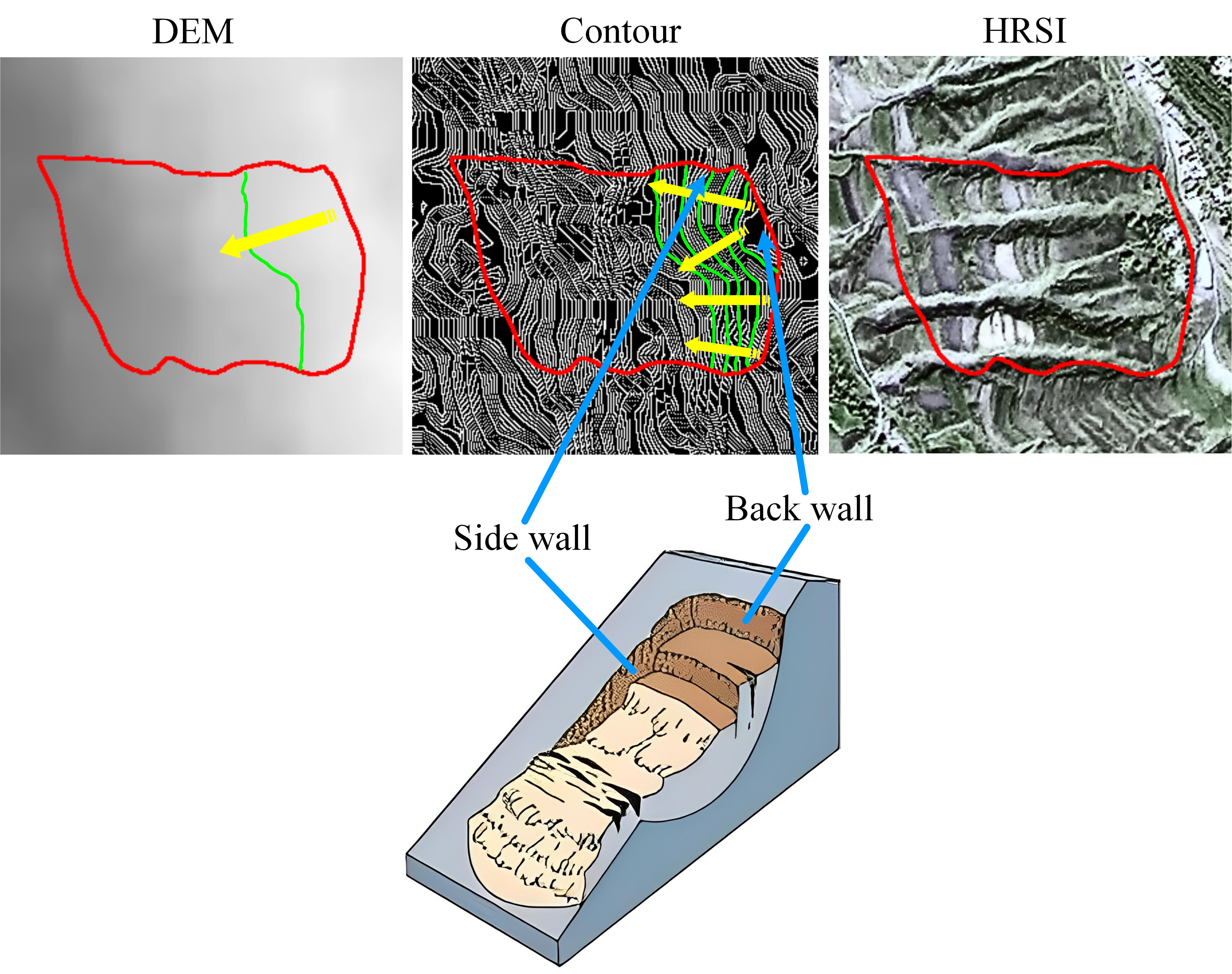}
    \DeclareGraphicsExtensions.
    \caption{Illustration of projections from a 3D landslide (bottom) into DEM (left), Contour (middle) and HRSI (right). One of the salient features of a landslide is the steep altitude descent of back and side walls, which is located in light gray areas with high elevation in DEM. The contour map is obtained by applying the Gauss-Laplace edge operator to DEM, where the back wall is parallel to the contour lines, while the side wall is vertical to the contour lines highlighted in green. The yellow arrows indicate landslide aspect. The gradient variation features in DEM are projected as the texture features in HRSI. The bottom schematic diagram of landslide is taken from \cite{weblandslide}.}
    \label{walls}
\end{figure}

In order to tackle the small-sized dataset problem, we employ contrastive learning to enhance feature extraction by distinguishing positive samples from negative ones in the feature space \citep{wu2018unsupervised, he2020momentum, chen2020simple}. Based on expert knowledge that the boundary features of back and side walls reflect the most elevation change of landslides as illustrated in Fig. \ref{walls}, we propose a supervised hyper-pixel-wise contrastive learning method forcing the model to learn these local salient features in the feature space with great efficiency and reliability. The hyper pixel, defined as an area of the original image corresponding to a pixel in the high-level feature map by position, contains much rich and abundant semantic information from a local region compared to the information captured by individual pixels in pixel-wise contrastive learning.

In general, our main contributions are summarized in the following:
	\begin{itemize}
		\item We propose the HPCL-Net model for relic landslide detection task to address the visual blur problem. Through designing a dual-branch Heterogeneous Feature Extractor with the Coordinate Attention mechanism, optical features such as texture and color in HRSIs and crucial topographic features such as altitude and gradient in DEM data are extracted and effectively fused. Results of ablation experiments and Visualization analysis in hot maps validate the effectiveness of our proposed model.
		
		\item We propose a supervised hyper-pixel-wise contrastive learning method to tackle the small-sized dataset problem. Under the guidance of experts knowledge in the landslide recognition, hyper-pixel-wise sample pairs are constructed from the areas of the back walls and side walls of a landslide to enhance the local salient feature extraction ability of the model. To significantly improve the effectiveness of HPCL, we build global category queues achieving cross-image contrast between anchors and keys in contrastive learning, and update the keys by momentum encoder to avoid the asynchronous update problem.
		
		\item Extensive experiments are conducted to evaluate the proposed model on a real relic landslide dataset, and the experimental results demonstrate that HPCL-Net achieves a significant improvement in the reliability of detecting relic landslides.
	\end{itemize}

\section{Data and preprocessing}
\subsection{Study area}
\begin{figure}[h]
    \centering
    \includegraphics[scale=0.6]{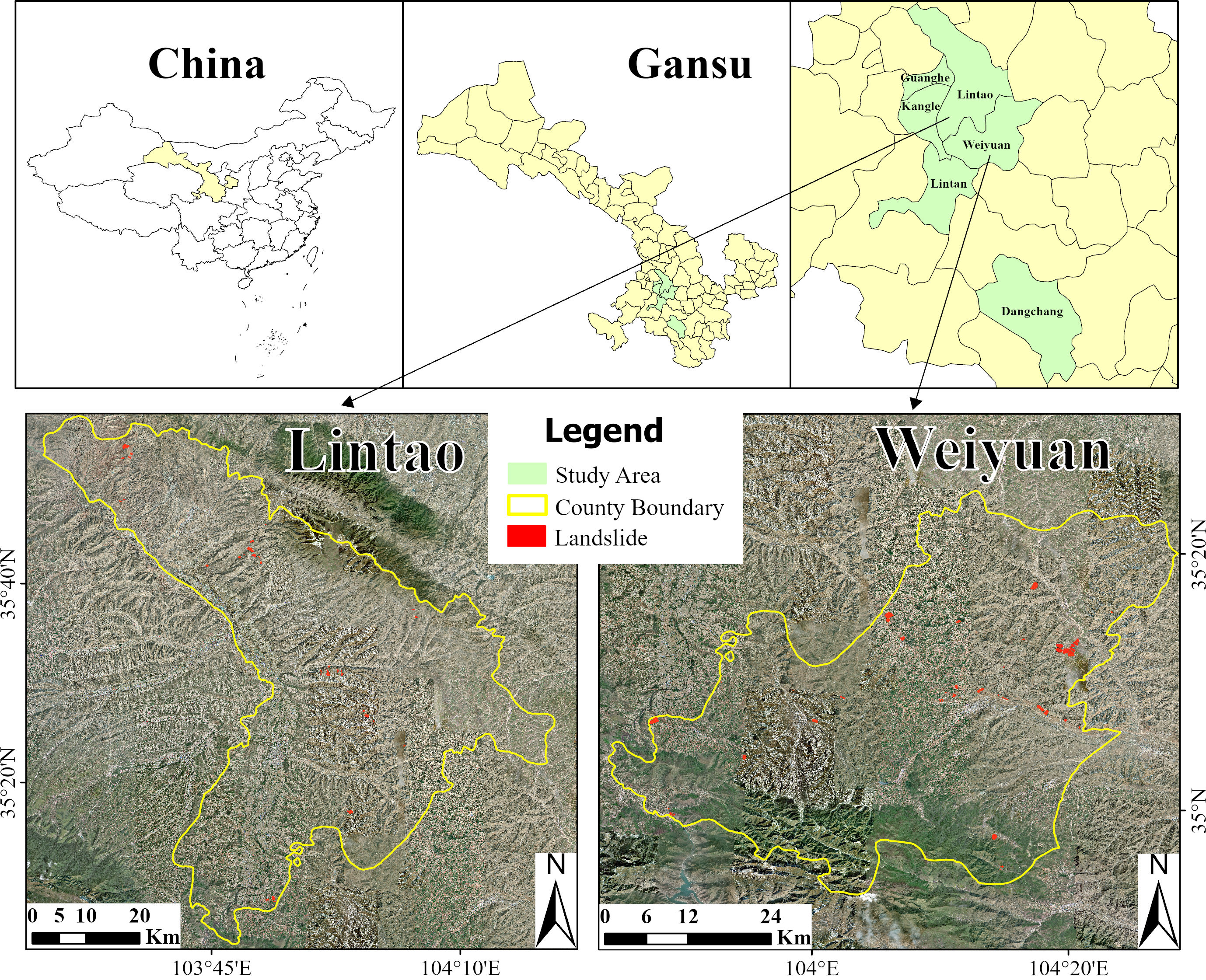}
    \DeclareGraphicsExtensions.
    \caption{The geo-location of the study area and the distribution maps from Lintao and Weiyuan.}
    \label{StudyArea}
\end{figure}

The study area locates in the Loess Plateau at the intersection of the western end of the Western Qinling Fold Belt and the southeastern end of the Qilian Fold Belt. Specifically, the relic landslides are distributed in six counties of Gansu Province of China, i.e., Kangle, Weiyuan, Lintan, Lintao, Guanghe, and Dangchang. Fig. \ref{StudyArea} shows the location of the study area and the landslide distribution maps from Lintao and Weiyuan. And the distribution maps of the remain four counties are presented in Appendix A(Fig. \ref{Areas}).

The low mountain and hilly landforms in the area are generally developed, with vertical and horizontal gullies. The stratigraphic lithologies are dominated by loose loess on the surface and weak mudstone in the lower part, with residual and alluvial parent material of soil. Besides, most of the soil moisture conditions in the area are not suitable for forest development, and the natural vegetation is grassland and shrub. Coniferous forest and deciduous broad-leaved forest are distributed in southern part of the study area since climate there is relatively humid \citep{CYKX20231218005}. Therefore, the vegetation coverage and the viscous soil with weak water permeability, corrosion resistance and erosion resistance are attributed to the the occurrence of landslides in the past.

The northern part of the study area is a cold temperate semiarid zone, and the southern part is a cold temperate subhumid zone, with typical continental monsoon climate characteristics \citep{ZMLC}. The climate in winter and summer is cold, dry, windy and sandy, affected by polar dry and cold air masses. And the climate in summer and autumn is hot, with frequent rainstorm, affected by the Western Pacific subtropical high and Indian Ocean depression. Under such climate conditions, the occurrence period of landslides in the study area is very concentrated, with the highest occurrence occurring from July to September each year, accounting for more than 68$\%$ of all landslides in the whole year. In addition, March is also the peak period for landslides in the area \citep{DHW}.

\subsection{Loess Plateau relic landslide dataset}
The dataset used in the paper were measured in 2018 by the Gaofen-1 satellite, and contains 168 relic landslide samples cropped to the size of $512 \times 512$. The HRSIs are RGB images and the DEM data are grayscale images. The images in the dataset are divided into the training, validation, and test sets in a ratio of 6:2:2. The resolution of the HRSI is 2m/pixel, and the resolution of the DEM data is originally 30m/pixel and subsequently interpolated to match the resolution of the HRSI. Since there are only 266 landslide samples, 5-fold cross validation is employed to achieve more stable experimental results, which differs from the fixed dataset partition scheme in the reference baseline model \citep{liu2023feature}. Both the dataset partition scheme and the extended dataset cause the greater difference in numerical results of the baseline model.

All the training and validation images are augmented by horizontal flipping, vertical flipping and rotating for $90^{\circ}$, $180^{\circ}$ and $270^{\circ}$. Histogram equalization is implemented as well.

\section{Methods}
\begin{figure*}[h]
    \centerline{\includegraphics[scale=0.7]{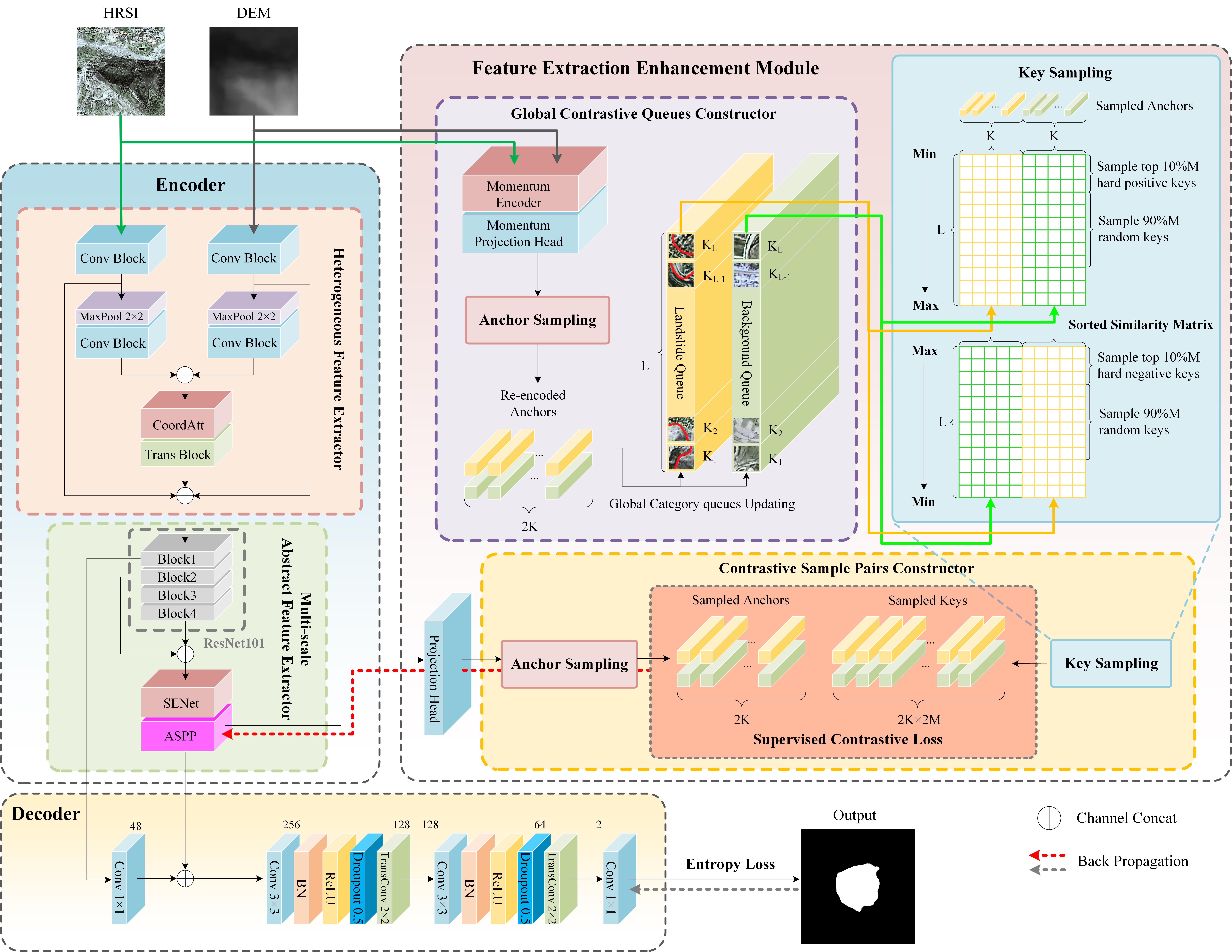}}
    \DeclareGraphicsExtensions.
    \caption{Architecture of HPCL-Net.}
    \label{IntegralStructure}
\end{figure*}

Considering the visual blur and small-sized dataset problems, we propose the HPCL-Net model using HRSI and DEM data for reliable detection of relic landslide through feature fusion in the semantic space and supervised contrastive learning. As shown in Fig. \ref{IntegralStructure}, the HPCL-Net has a typical encoder-decoder architecture. Specifically, the encoder first extracts optical features from HRSIs and topographical features from DEM data separately, and fuses them in the semantic space by a two-branched Heterogeneous Feature Extractor (HFE) module. Then a Multi-scale Abstract Feature Extractor (MAFE) module in the encoder further extracts higher-level semantic features, from which the Decoder classifies each pixel into landslide or background. Besides, a Feature Extraction Enhancement (FEE) module based on supervised hyper-pixel-wise contrastive learning can help the encoder to distill the essential differences between the semantic features of landslides and background slopes, and learn the salient features of landslides.

\subsection{Encoder}
As shown in Fig. \ref{IntegralStructure}, the encoder consists of HFE and MAFE. Improved from our previous work \citep{liu2023feature}, the newly designed encoder takes DEM data as an independent channel, and the heterogeneous features extracted independently from HRSIs and DEM data are fused through channel concatenation with the CA mechanism, contrasted to the scheme in \cite{liu2023feature} which duplicates the DEM data three times to facilitate the pixel addition with the RGB channel.

\subsubsection{Heterogeneous Feature Extractor}
\begin{figure}[h]
    \centerline{\includegraphics[scale=1.3]{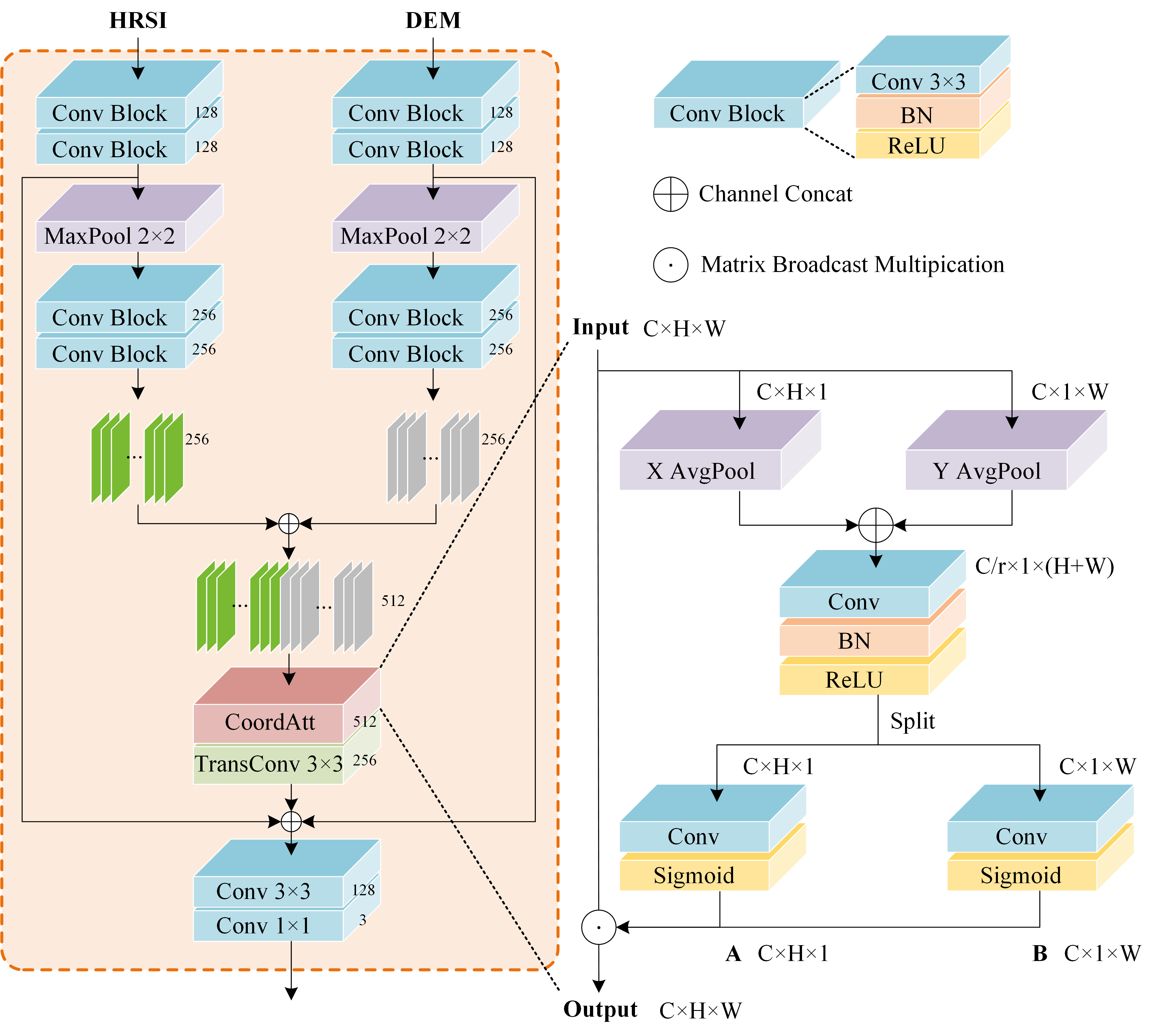}}
    \DeclareGraphicsExtensions.
    \caption{Architecture of Heterogeneous Feature Extractor. The structure of Coordinate Attention mechanism is indicated by the dotted line and the number on the right side of each component indicates the number of output channels.}
    \label{Siamese}
\end{figure}

In order to distill features independently from HRSI and DEM, HFE module is designed as a dual-branch network followed by a feature fusion network, whose architecture is detailed in Fig. \ref{Siamese}. The two branches hold the same architecture but do not share weights, consisting of a Conv Block, a max pooling layer, and a Conv Block sequentially, where the Conv Block includes a convolution layer, a batch normalization layer, and a ReLU layer. The HRSIs and DEM data are fed into the dual-branch network to first obtain two feature maps that are fused via channel concatenation and then weighted via the Coordinate Attention (CA) mechanism.

As shown in Fig. \ref{Siamese}, through the CA mechanism, the heterogeneous information are fused, and the direction and position-sensitive information are distilled, allowing the model to locate landslides more accurately. The weighted feature map is upsampled and concatenated with the feature maps produced by the first Conv Block, and the subsequent CNN extracts both high-level abstract semantic information and low-level spatial detailed syntactic information. In HFE, batch normalization (BN) and ReLU are utilized after each layer to increase model stability and avoid overfitting.

\subsubsection{Multi-scale Abstract Feature Extractor}
In order to distinguish relic landslides from backgrounds with high similarity, much higher abstract semantic features are required to extract from the fused features, which is implemented by MAFE. The backbone of the module is the dilated ResNet101 \citep{chen2018encoder} that has been widely proven to have an excellent feature extraction capability. The ResNet101 consists of four dilated residual blocks. In order to retain more detailed features, the feature map of Block2 is concatenated with the upsampled feature map of Block4. The concatenated feature map is adaptively weighted through the SENet \citep{hu2018squeeze}. Then, the Atrous Spatial Pyramid Pooling (ASPP) \citep{7913730} is used to effectively extract features at different scales through multiple parallel atrous convolution with different dilation rates. In addition, the global average pooling is used to gain global context information of the feature map.

\subsection{Feature Extraction Enhancement Module}
FEE is based on designed supervised hyper-pixel-wise contrastive learning to extract local salient features from the boundaries of landslides, which consists of two components, i.e., the Contrastive Sample Pairs Constructor (CSPC) and the Global category Queues Constructor (GCQC), where CSPC constructs hyper-pixel-wise positive and negative sample pairs, and GCQC enhances the efficiency and performance of contrastive learning. The architecture of the module is shown in Fig. \ref{IntegralStructure}.

\begin{figure}[h]
    \centerline{\includegraphics[scale=1]{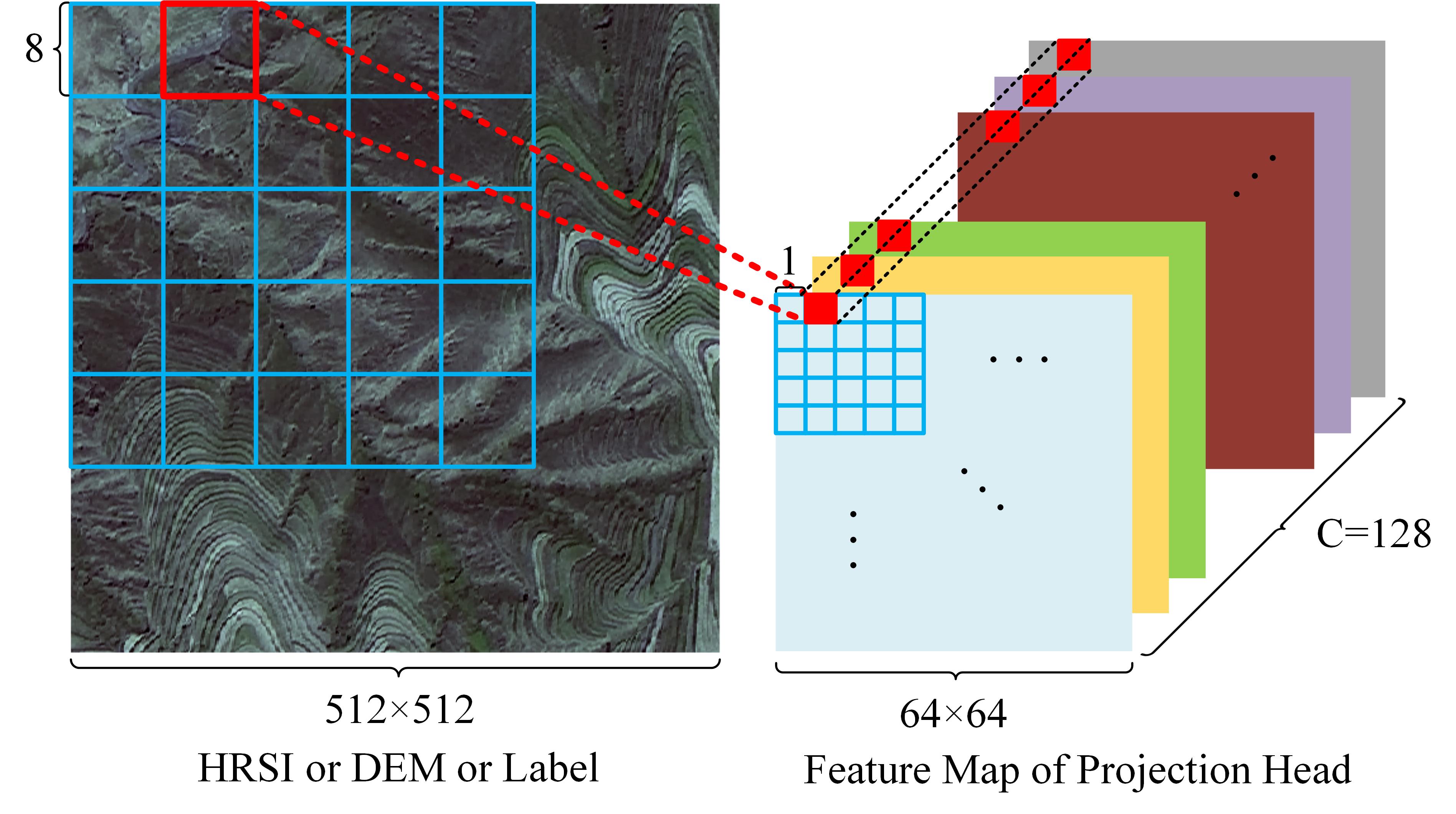}}
    \DeclareGraphicsExtensions.
    \caption{Illustration of the hyper-pixel definition. Each $8 \times 8$ blue square on HRSI, DEM or Label is a hyper-pixel corresponding to a pixel of 128 channels in the feature map produced by the projection head.}
    \label{HyperPixel}
\end{figure}

Since the essential features of landslides are the gradient pattern, it is crucial to find the boundary areas of the back and side walls where the elevation is significantly decreased. For CSPC, under the anchor sampling strategy, the pixels from the high-dimensional feature map of the projection head are sampled as anchors. Each pixel corresponds to a hyper-pixel that is located at the boundary areas of the back walls/side walls or the background area. And a hyper-pixel consists of a patch of pixels in the original image far smaller than landslides. As illustrated in Fig. \ref{HyperPixel}, the patches are determined by label and DEM. Under the key sampling strategy, keys are selected from the queues to construct positive and negative sample pairs with anchors. Then, the supervised contrastive loss is calculated by the sample pairs, where `supervised' suggests that the class of each anchor and key is known from label.

In CGQC, the global category queues contain a landslide queue and a background queue, and both queues are stored outside the model. The queues are updated on-the-fly through enqueue and dequeue operations, where the enqueue operation denotes pulling the anchors re-encoded by the momentum encoder from the latest mini-batch into the queues, and the dequeue operation denotes pushing the keys from the oldest mini-batch out of the queues. Moreover, keys in the queues are selected from the entire dataset, which facilitates global contrastive learning for rich sample diversity.

\subsubsection{Contrastive Sample Pairs Constructor}
Hyper-pixel-wise anchors and keys are constructed and encoded to build sample pairs by CSPC, and the three main processings are anchor sampling, key sampling, and supervised contrastive loss calculation.

As for anchor sampling, in each batch $K$ landslide anchors and $K$ background anchors are sampled from the high-level feature map produced by MAFE of the encoder, and compared with the keys selected from the global category queues. The landslide anchors are the hyper pixels located in the back and side walls, and the background anchors are hyper pixels randomly sampled from the background of input images.

\begin{figure}[htbp]
  \centerline{\includegraphics[scale=1.2]{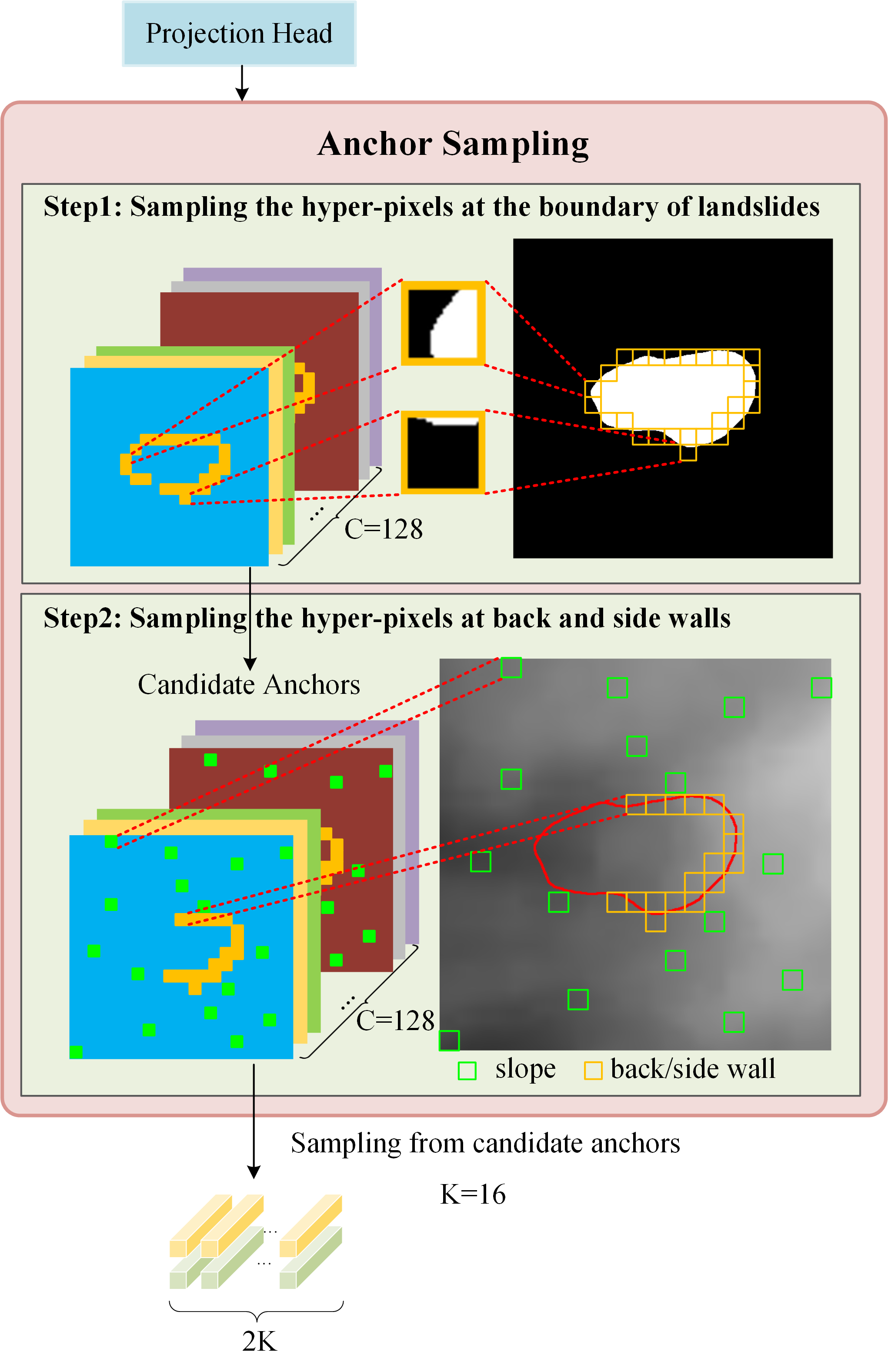}}
  \DeclareGraphicsExtensions. 
  \caption{Anchor Sampling schematics. On the left of Steps 1 and 2 is the feature map produced by the projection head, from which anchors are sampled. Yellow boxes and squares mark sampled landslide hyper-pixels and anchors, while green boxes and squares mark sampled background hyper-pixels and anchors. The correspondence between hyper-pixels and anchors is shown by the red dashed line.}
  \label{AnchorSampling}
\end{figure}

Since the sizes of the HRSI, DEM and label are $512 \times 512$, and the high-level feature maps are reduced to $64 \times 64$ through a series of convolution and down-sampling operations, each pixel in the feature maps corresponds to a hyper-pixel which is an $8 \times 8$ patch in the HRSI, DEM and label by location. If a hyper-pixel is located in the boundary area of back wall/side wall, the corresponding pixel is then sampled from the feature map as a candidate landslide anchor. The position of the candidate landslide hyper-pixel is determined in two steps as illustrated in Fig. \ref{AnchorSampling}, i.e., (1) The hyper-pixels with more than 6 and less than 58 landslide pixels are determined, which correspond to the boundary areas of the landslides; (2) Among these hyper-pixels, those with an average elevation greater than the median elevation of the entire landslide according to DEM are sampled candidate landslide anchors, meaning they are located on the back and side walls. The hyper-pixels without any landslide pixels within are sampled as candidate background anchors.

The number of the channels in the high-level feature map from the encoder is 256 and then reduced to 128 through the projection head, so the anchor is a 128-dimensional vector. $K$ anchors are randomly sampled from the candidate anchors. We set $K$ to be 16.

As for key sampling, for each of the aforementioned 2$K$ anchors, $M$ keys from the queue of the same category as the anchor are selected to construct the positive sample pairs with the anchor, another $M$ keys from the queue of the different category are selected to construct the negative sample pairs with the anchor, according to the similarities between anchors and keys.

In practice, the similarities between the anchor and all keys in queues are first calculated. In the queue of the same category, the top 10\% $M$ keys with the lowest similarities are selected as the hard positive samples. Similarly, In the queue of the different category, the top 10\% $M$ keys with the highest similarities are selected as the hard negative samples. The process of key sampling is illustrated in Fig. \ref{IntegralStructure}. These two types of hard samples can contribute more to the improvement of the discriminative ability \citep{khosla2020supervised, robinson2020contrastive}, and the remaining 90\% $M$ keys are randomly sampled from the remaining keys in the queues, where $M$ is set to be 1000.

The positive and negative sample pairs are used to calculate the supervised contrastive loss defined as follows

\begin{equation}
  L_{i}^{SC} = \frac{1}{\left| P_{i} \right|}{\sum\limits_{i^{+} \in P_{i}}{- {\rm log}\frac{{\rm exp}\left( i \cdot i^{+}/\tau \right)}{{\rm exp}\left( i \cdot i^{+}/\tau \right) + {\sum_{i^{-} \in N_{i}}{{\rm exp}\left( i \cdot i^{-}/\tau \right)}}}}},
\end{equation}
where $i$ denotes anchor $i$ sampled in the current batch, $P_{i}$ and $N_{i}$ denote the $M$ positive and negative samples, $i^{+}$ and $i^{-}$ denotes a key in $P_{i}$ and $N_{i}$ respectively, $\left| P_{i} \right|$ denotes the total number of $P_{i}$, i.e., $M$. $\tau$ is the temperature for controlling the shape of the distribution of logits that is in logarithmic scale. As $\tau$ decreases, the distribution of logits becomes more peak, enabling the model to focus on the hard samples. While as $\tau$ increases, the distribution of logits becomes smoother, reducing the sensitivity of the model to the difference between the anchors and keys. $\tau$ is set to 0.1 here. In addition, the larger $i \cdot i^{+}$ and the smaller $i \cdot i^{-}$, the smaller $L_{i}^{NCE}$ is. Therefore, by optimizing the loss, the distance between the anchors and positive keys will be closer and the distance between the anchors and negative keys will be farther.

\subsubsection{Global Category Queues Constructor}
GCQC is designed to satisfy the requirement for abundant sample pairs and improve the effectiveness of contrastive learning, where the specific configurations of the global category queues, momentum encoder and projection head are elaborated on below.

The global category queues include the landslide and the background queues of length $L$. Since keys are vectors of dimension $D$, the queue is a tensor of size $L \times D$. In addition, the queue adopts the first-in-last-out update strategy: $K$ new keys from latest mini-batch go into the queue, and $K$ old keys from the oldest mini-batch go out of the queue. Accordingly, $K$ is independent of $L$. By setting appropriate values of $K$ and $L$, substantial computing resources are saved and the training efficiency is thus improved. The queues are randomly initialized and a momentum encoder is used to re-encode anchors as new keys to update queues.
  
The momentum encoder is introduced in GCQC to allows the new keys that are enqueued in different batches to be encoded by similar encoder parameters, avoiding the asynchronous update issue problem. The projection head that consists of a simple $1 \times 1$ convolutional layer is also introduced to reduce feature dimension for the improvement of training efficiency.
  
Practically, the momentum encoder and momentum projection head have the same architectures as the encoder and normal projection head, and are updated through momentum-based moving average shown below rather than back propagation. By feeding HRSIs and DEM data to both the encoder and momentum encoder, the pixels in the same positions as the anchors are sampled from the high dimensional feature map from the momentum projection head. These pixels are regarded as the re-encoded anchors, and then dimensionally reduced and pushed into the queues.
\begin{equation}\theta_{k}^{n} \gets \left(\theta_{k}^{n-1} + \theta_{q}^{n} \right) \,,
\end{equation}
\begin{equation}P_{k}^{n} \gets \left(P_{k}^{n-1} + P_{q}^{n} \right) \,,
\end{equation}where $\theta_{k}^{n}$, $\theta_{q}^{n}$, $P_{k}^{n}$, and $P_{q}^{n}$ denote the parameters of the encoder, the momentum encoder, the projection head, and the momentum projection head, respectively, from mini-batch $n$. $\theta_{k}^{n-1}$ and $P_{k}^{n-1}$ denote the parameters of the momentum encoder and the momentum projection head from mini-batch $n-1$, respectively, $m$ denotes momentum and is set to 0.999 here.

\subsection{Decoder}
Decoder is built to recover the resolution of the high-level feature maps to input size, and output the category score maps to obtain segmentation results. The architecture of the decoder is shown in Fig. \ref{IntegralStructure}. The feature maps of Block1 in the ResNet101 are concatenated with the output of MAFE in order to integrate low-level features such as details and spatial features with high-level abstract features, and then fed into the decoder. The trainable filters of the transposed convolution reduce redundant information and have a better feature mapping capability to reconstruct the resolution of the input images.

The Cross-Entropy loss formulated below is used to calculate the loss between the category score map and the label where the landslide pixels are marked as 1 and others as 0.
\begin{equation}
    L_{i}^{CE} = - y_{i}\left. {\mathit{\log}(}p_{i} \right) + \left( 1 - y_{i} \right)\left. {\mathit{\log}(}1 - p_{i} \right),
\end{equation}
where $y_{i}$ denotes the ground truth of pixel $i$ in the categorical score map and $p_{i}$ indicates the probability of pixel $i$ being classified into landslide.

Therefore, the total training loss is
\begin{equation}
    L = {\sum_{i}} {\left(\alpha L_{i}^{CE} + \beta L_{i}^{SC}\right)},
\end{equation}
where $\alpha$ and $\beta$ are set to 1.0 and 0.1, respectively.

\subsection{Model evaluation}
\subsubsection{Reference model}
We compare the HPCL-Net with the FFS-Net \citep{liu2023feature} that is considered as the baseline model. Since the FFS-Net has shown that it outperforms universal semantic segmentation models such as the UNet and Deeplabv3+ for landslide detection, we do not need to repeat the same comparison experiments. In addition, other multi-modal data fusion models \citep{mondini2019sentinel, ghorbanzadeh2019evaluation, DSMFusion2019, ji2020landslide, MultiModal2019, Zeng2020} use the data such as DSM, multi-spectral bands, aerial photograph and InSAR greatly different from our HRSIs and DEM data, where DSM and aerial photograph have much higher resolution, and InSAR contains distinctive semantic features such as deformation rate. So, it is difficult to directly migrate the HRSIs and DEM data to these models. Besides, the tasks of these models are different from ours, e.g., the landslide classification \citep{ji2020landslide} and image registration \citep{Zeng2020}, and the source codes of these models have not been publicly released, so we cannot reproduce their results. Considering the above cases, we have to solely compare our model with the FFS-Net.

In addition, we argue that the simple pixel-wise addition performs well in \cite{liu2023feature} because the very limited number of samples cannot support high-capacity encoder. However, when data is fully utilized in a more efficient way, a more powerful encoder is expected to enhance the feature extraction performance. Therefore, on the basis of the designed contrastive learning scheme that enables a much stronger feature extraction ability on small-sized datasets, we upgrade the encoder in \cite{liu2023feature} through channel concatenation and CA mechanism \citep{hou2021coordinate} when fusing heterogeneous features, so that the contrastive learning and improved encoder can bring the best in each other for efficient and reliable semantic feature extraction.

\subsubsection{Hyperparameters}
In all experiments, the stochastic gradient descent is chosen as the optimizer and the polynomial annealing policy \citep{chen2017rethinking} is used to schedule the learning rate. The hyperparameters and their values in this study are shown in Table~\ref{Hyperparameters}.
\begin{table}[h]
  \centering
  \caption{Hyperparameters.}
  \label{Hyperparameters}
  \begin{tabular}{ll}
\hline
Item & Value\\
\hline
Mini-batch Size & 2\\
Epoch & 100\\
Initial learning rate & 0.007\\
Weight decay & 0.007\\
Momentum & 0.9\\
Optimizer & SGD\\
\hline
  \end{tabular}%
\end{table}
\subsubsection{Performance metrics}
Following \cite{chen2017rethinking, 7913730, wang2021exploring, chen2018encoder, liu2023feature}, the mainstream performance metrics such as $Precision$, $Recall$, $F1score$, $Landslide\_IoU$ and mean Intersection over Union ($mIoU$) are employed, which are defined as follows
\begin{equation}
    precision = \frac{TP}{TP + FP}\,,
\end{equation}
\begin{equation}
    recall = \frac{TP}{TP + FN}\,,
\end{equation}
\begin{equation}
    F1score = 2 \times \frac{precision \times recall}{precision + recall}\,,
\end{equation}
\begin{equation}
    Landslide \underline{~} IoU = \frac{TP}{TP + FP + FN}\,,
\end{equation}
\begin{equation}
    Background \underline{~} IoU = \frac{TN}{TN + FN + FP}\,,
\end{equation}
\begin{equation}
    mIoU = \frac{1}{2} \times (Landslide \underline{~} IoU + Background \underline{~} IoU)\,,
\end{equation}
where $TP$, $TN$, $FP$, $FN$ denote the numbers of correctly predicted landslide pixels, correctly predicted non-landslide pixels, incorrectly predicted landslide pixels, and incorrectly predicted non-landslide pixels, respectively.

\section{Results and Discussion}
In this section, we carry out comparison experiments, ablation experiments, and cross validation experiments, along with discussions on the experimental results. To better illustrate the training process of HPCL-Net, it is shown in the form of pseudocode in Appendix B.

\subsection{Overall comparative experiments}
The numeric results in Table~\ref{FFS-Net2ours} show that our proposed HPCL-Net performs better than the FFS-Net with $mIoU$ increased by 3.1\%, $Landslide\_IoU$ increased by 6.0\%, and $F1score$ increased by 6.4\%, indicating reliable detection of relic landslides. These improvements validate the effectiveness of the contributions in this paper.
\begin{table}[htbp]
    \centering
    \caption{Numerical results of segmentation networks.}
    \begin{tabular}{lccccc} 
        \hline
        \multicolumn{1}{l}{Method}& Precision & Recall & F1 & Landslide\_IoU & mIoU\\
        \hline
        FFS-Net & 0.462 & 0.551 & 0.503 & 0.334 & 0.620\\
        HPCL-Net & \textbf{0.579} & \textbf{0.573} & \textbf{0.576} &  \textbf{0.394} & \textbf{0.651}\\
        \hline
    \end{tabular}
    \label{FFS-Net2ours}
\end{table}

\begin{figure*}[h]
    \centerline{\includegraphics[scale=0.4]{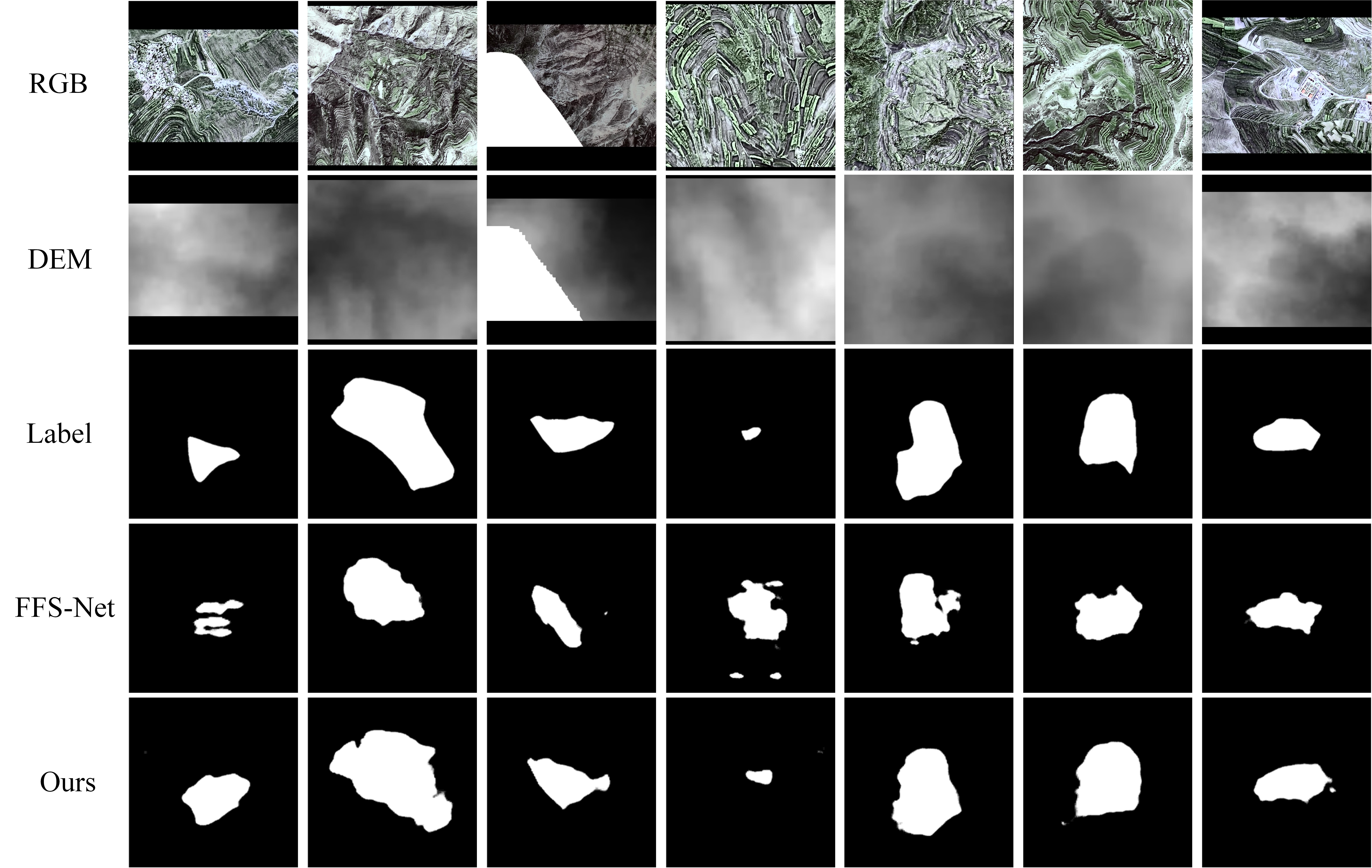}}
    \DeclareGraphicsExtensions.
    \caption{Graphical segmentation results on the relic landslides. From top to bottom are RGB, DEM, label, FFS-Net and ours.}
    \label{ComResult}
\end{figure*}

The graphical segmentation results in Fig. \ref{ComResult} show that, compared to the FFS-Net, our results retain finer boundaries and are more accurate in shape and position, which are consistent with the numerical results.

\begin{figure*}[h]
    \centerline{\includegraphics[scale=0.45]{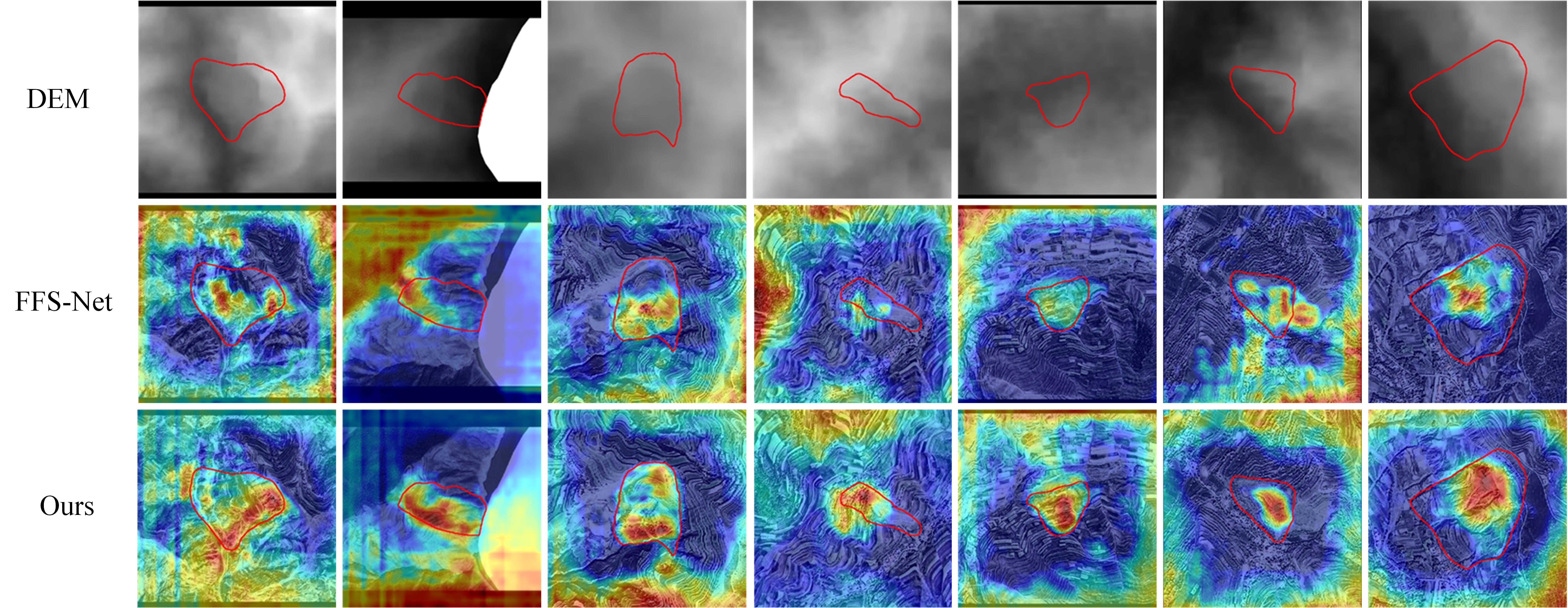}}
    \DeclareGraphicsExtensions.
    \caption{Visualization results for the last layer of the encoder by Grad-CAM. The landslides are outlined by the red line.}
    \label{ComHeat}
\end{figure*}

To visualize the landslide features learned by each model, the Gradient-weighted Class Activation Mapping (Grad-CAM) scheme \citep{selvaraju2017grad} is employed to produce a heatmap, where features with more contribution to the target class have higher heat. We apply Grad-CAM to the last layer of the encoders of the FFS-Net and HPCL-Net, and the heatmaps are shown in Fig. \ref{ComHeat}. Compared to the FFS-Net, our model has more features with higher heat located in the back and side walls of the landslides, proving that the proposed model is capable of effectively learning the crucial features of relic landslides.

\subsection{Ablation experiments}
To precisely evaluate the contribution of each module in the proposed model, we conduct ablation experiments. The numerical results are shown in Table~\ref{ablation}, where FFS-Net is regarded as baseline; `B+F' denotes employing newly designed HFE on the baseline; `B+F+C' denotes a simplified end-to-end version of the HPCL-Net where the anchors and keys for contrastive learning are only sampled within the current mini-batch; `B+F+C+G' denotes the final version of the HPCL-Net. 

The numerical ablation results show that firstly by improving the feature extraction and fusion module in the FFS-Net, the model performance is enhanced, where $mIoU$ is increased by 0.5\%, $Landslide\_IoU$ is increased by 0.5\%, and $F1score$ is increased by 0.5\%. These enhancements show the effectiveness of designed HFE. Secondly, after adding CSPC, although $mIoU$ is slightly decreased, $Landslide\_IoU$, which is more essential for the task in practice, is increased by 1.2\%, and the $F1score$ is also increased by 1.4\% from the previous experiment. These enhancements demonstrate that the proposed CSPC makes certain contributions to the identification of landslide features. As for the reasons of no increase in $mIoU$, we argue that the end-to-end training paradigm is difficult to satisfy the necessary requirements for a large number of samples in contrastive learning. So we design GCQC to fix this problem. Finally, after applying GCQC to provide sufficient samples and improve sample diversity, the performance of the model is significantly improved, where $mIoU$ is increased by 2.7\%, $Landslide\_IoU$ is increased by 4.3\%, and $F1score$ is increased by 4.5\%, proving our previous argument and the effectiveness of GCQC.

\begin{table}[h]
    \centering
    \caption{Numeric results of ablation experiments.}
    \begin{tabular}{lcccccr}
  	\cmidrule{1-6}  Scheme & Precision & Recall & F1 & Landslide\_IoU & mIoU  &  \\
  	\cmidrule{1-6}
	\cmidrule{1-6}  B     & 0.462 & 0.551 & 0.503 & 0.334 & 0.620  &  \\
					B+F   & 0.449 & 0.589 & 0.510 & 0.339 & 0.625 &  \\
					B+F+C & 0.545 & 0.520  & 0.532  & 0.351 & 0.624 &  \\
					B+F+C+G & 0.579 & 0.573 & 0.576 & 0.394 & 0.651 &  \\
	\cmidrule{1-6}    
	\multicolumn{6}{l}{B: Base model, i.e., FFS-Net} &  \\
	  \multicolumn{6}{l}{F: Heterogeneous Feature Extractor} &  \\
	  \multicolumn{6}{l}{C: Contrastive Sample Pairs Constructor} &  \\
	  \multicolumn{6}{l}{G: Global Category Queues constructor} &  \\
  	\cmidrule{1-6}    
    \end{tabular}
    \label{ablation}
  \end{table}

\begin{figure}[h]
    \centerline{\includegraphics[scale=0.57]{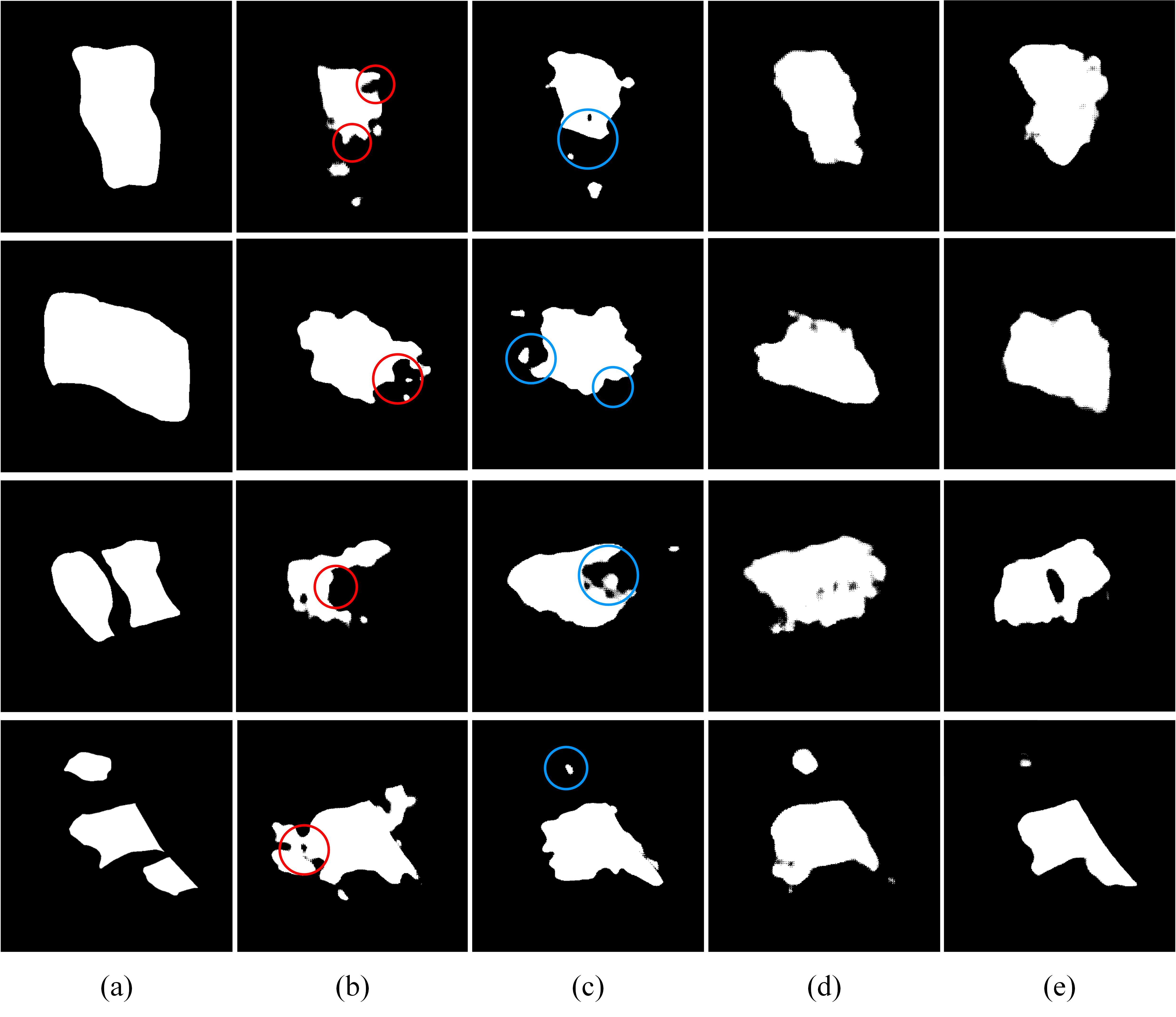}}
    \DeclareGraphicsExtensions.
    \caption{Visualization results for the encoder feature map. (a) label, (b) B, (c) B+F, (d) B+F+C, (e) B+F+C+G.}
    \label{AblResult}
\end{figure}

The graphical segmentation results of the four ablation experiments are shown in Fig. \ref{AblResult}. HFE can reduce misclassifications of the landslides, as highlighted by the red circles. After adding CSPC, the shape of the segmented landslide is more complete, and the position is more accurate, as highlighted by the blue circles. For the final model, the segmentation of the boundaries are the most clear and precise, which is consistent with the numerical results.

\begin{figure}[h]
    \centerline{\includegraphics[scale=0.75]{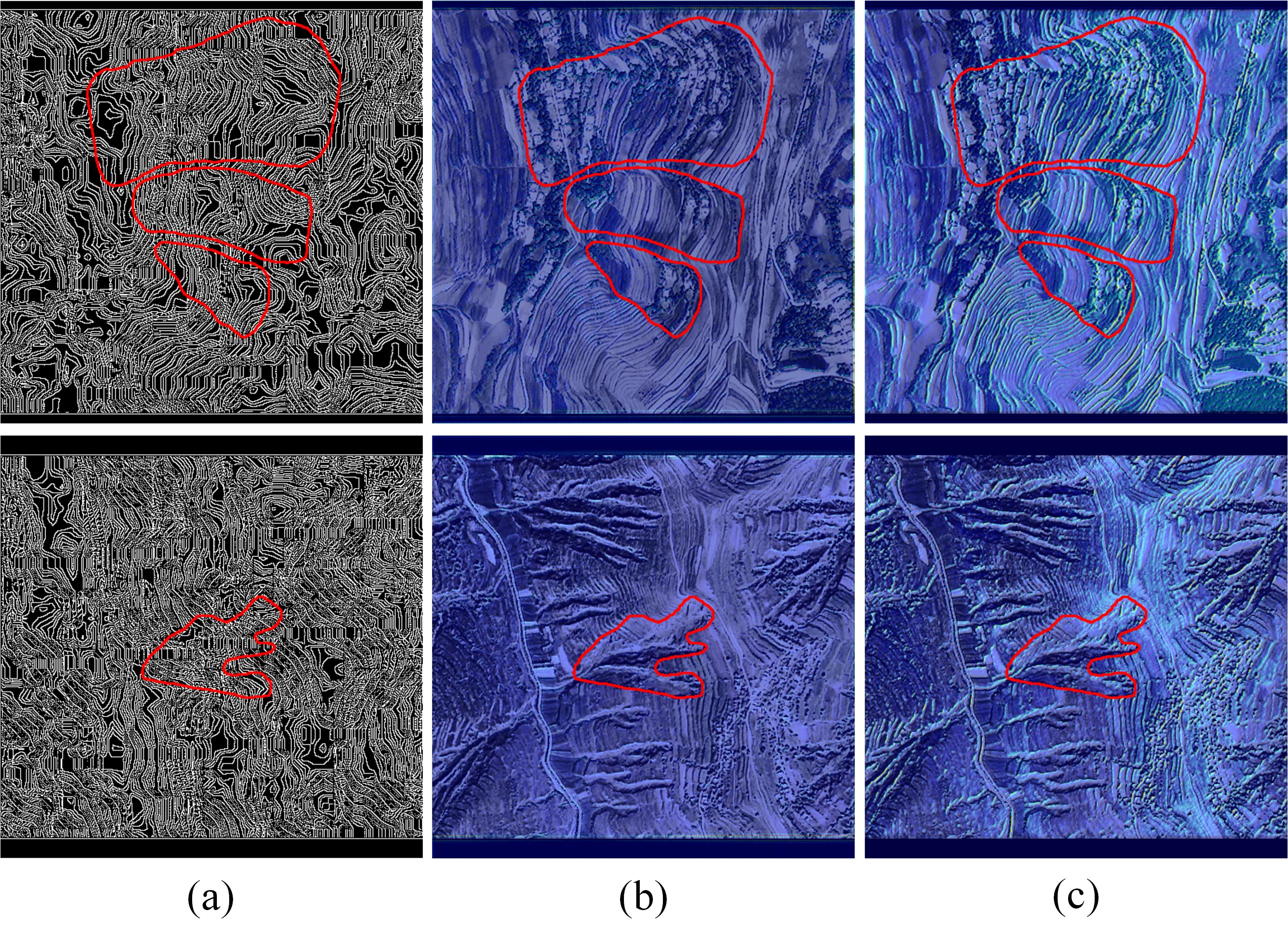}} 
    \DeclareGraphicsExtensions.
    \caption{Visualization results for the last layer of the feature extraction and fusion module in FFS-Net and Heterogeneous Feature Extractor in HPCL-Net. The outline of the target landslide is highlighted in red line. (a) contour map, (b) B, (c) B+F.}
    \label{AblHeat1}
\end{figure}

To visualize the enhancements of the learned features corresponding to the increases of the performance metrics, we first choose the last layer of the feature extraction and fusion module in `B' and the heterogeneous feature extractor in `B+F' as the target layer of the Grad-CAM. The heatmaps are shown in the Fig. \ref{AblHeat1}, where the images in the last column are the contour maps obtained by applying the Gauss-Laplace edge operator to the DEM data. As can be seen from the figure, the heatmaps from `B' have almost no visible optical or topographic features, while the heatmaps from `B+F' have higher heat at the positions where elevation changes, forming the texture similar to the contour map. This indicates that by changing the fusion method to channel concatenation and introducing the CA mechanism, designed HFE can effectively integrate the gradient features from the DEM data and color, texture, and other optical features from the HRSI.

\begin{figure}[h]
    \centerline{\includegraphics[scale=0.22]{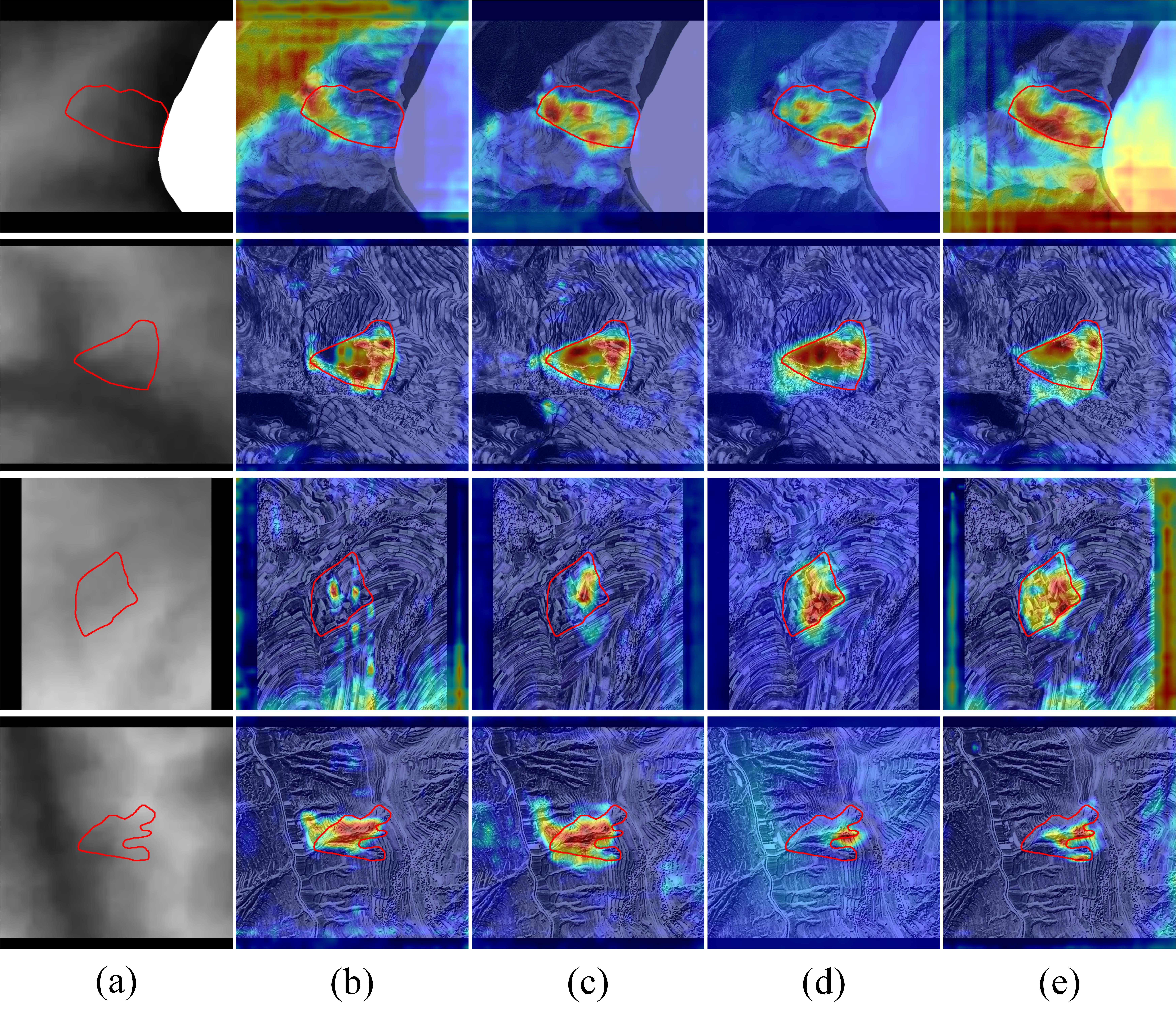}} 
    \DeclareGraphicsExtensions.
    \caption{Visualization results for the last layer of encoder by Grad-CAM. The outlines of landslides are highlighted by red line. (a) label, (b) B, (c) B+F, (d) B+F+C, (e) B+F+C+G.}
    \label{AblHeat2}
\end{figure}

Then, we choose the last layer in the encoders from the four experiments as the target layer of the Grad-CAM. Fig. \ref{AblHeat2} shows that the heatmaps generated by `B' and `B+F' have higher heat in the center of the landslide. However, after adding CSPC, the higher-heat areas in `B+F+C' tend to the back wall and side wall areas of the landslide, demonstrating that CSPC enables the model to focus on identifying essential features of landslides. In the heatmaps generated by `B+F+C+G', the higher-heat areas are mostly concentrated on the back wall and side wall areas, suggesting that GCQC, which facilitates training with sufficient and diverse samples, plays a significant role in enbling the model to effectively learn crucial features.

\subsection{Cross validation experiments}
In the section, we conduct cross-validation experiments for optimizing the model design. The baseline model is the HPCL-Net with the CA mechanism and the fusion method of channel concatenation. We first replace the channel concatenation with pixel addition. Then we replace the CA mechanism with the SENet and CBAM in the following experiments to compare the CA mechanism with the other two attention mechanisms.

The numerical results of the experiments are listed in Table~\ref{crossval}. The result of replacing channel concatenation demonstrates that the channel concatenation method is more appropriate than pixel addition to fuse the features with a significant semantic difference from HRSI and DEM, which is consistent with the previous analysis. The results of relacing the CA mechanism show that the CA mechanism enables the model to fuse features from both the channel and spatial dimensions and performs better than the SENet that focuses on only channel attention. Compared with CBAM that also focuses on channel and spatial attention, although $mIoU$ is almost the same, $F1score$ of the HPCL-Net is 1.0\% higher than the model with CBAM, proving the superiority of the CA mechanism for the landslide detection task.

\begin{table}[htbp]
    \centering
    \caption{Cross validation of fusion method.}
    \begin{tabular}{lccccc}
	\hline
	\multicolumn{1}{l}{Experiment}& Precision & Recall & F1 & Landslide\_IoU & mIoU\\
	\hline
	Baseline: CA + Channel Concat & 0.579 & 0.573 & 0.576 & 0.394 & 0.651\\
	CA + Pixel Addition & 0.583 & 0.531 & 0.556 & 0.374 & 0.638\\
	SENet + Channel Concat & 0.523 & 0.556 & 0.539 & 0.368 & 0.638\\
	CBAM + Channel Concat & 0.585 & 0.549 & 0.566 & 0.395 & 0.651\\
	\hline
	\end{tabular}
    \label{crossval}
\end{table}

Since the Kriging interpolating process on DEM data changes the real terrain distribution, we evaluate two schemes that differ in the way to exploit DEM data. The first scheme is interpolating the DEM data with Kriging interpolation algorithm to match the resolution of HRSI data, the other one is directly distilling semantic features from original DEM data without interpolation. The experimental results are listed in Table~\ref{interpolation1}, from which we can observe that the first scheme outperforms the second one in almost all measured performance metrics except the slide precision is slightly worse. 

\begin{figure}[htbp]
    \centerline{\includegraphics[scale=0.38]{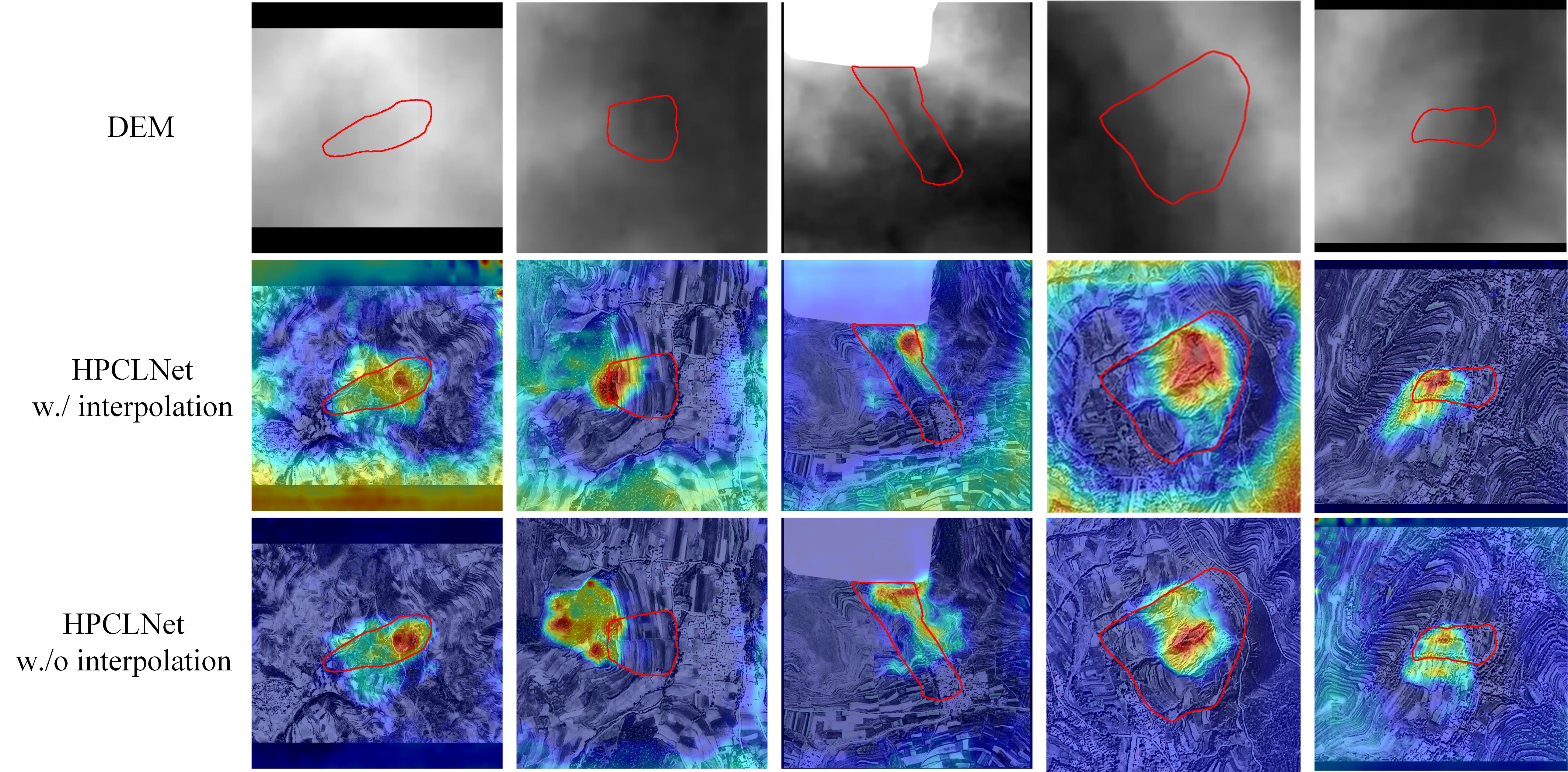}} 
    \DeclareGraphicsExtensions.
    \caption{Visualization results of two DEM distillation schemes.}
    \label{interpolation2}
\end{figure}

\begin{table}[htbp]
    \centering
    \caption{Numeric results of two DEM distillation schemes.}
    \begin{tabular}{lcccccr}
  	\cmidrule{1-6}  Scheme & Precision & Recall & F1 & Landslide\_IoU & mIoU  &  \\
  	\cmidrule{1-6}
	\cmidrule{1-6}  FFS-Net     & 0.462 & 0.551 & 0.503 & 0.334 & 0.620  &  \\
					HPCLNet w./ interpolation   & 0.585 & 0.549 & 0.566 & 0.395 & 0.651  \\
					HPCLNet w./o interpolation & 0.581 & 0.533  & 0.556  & 0.380 & 0.641 &  \\
	\cmidrule{1-6}    
    \end{tabular}
    \label{interpolation1}
  \end{table}
We also apply Grad-CAM to the last layer of the encoders, and the heatmaps are shown in Fig. \ref{interpolation2}. Although HPCLNet without interpolation tends to show higher heat located in the back and side walls of the landslides, the heat positions are less precise than that in HPCLNet with interpolation. 

As a result, the information presented by DEM data is the gradient pattern of both the back and side walls of landslides in the contour maps, which are one of the key features to recognize a landslide due to the steep altitude descent at back and side walls. In the case of gradient pattern recognition, the suitable resolution of DEM data might be different from HRSI data, and the cross validation experiment above verifies that even the resolution of the DEM data we used is low, it can still benefit the landslide detection. Besides, employing Kriging interpolating pre-processing method can present better landslide detection performance.

\subsection{Complexity analysis}
We also compare the complexity of the FFS-Net and HPCL-Net. The number of parameters and GFLOPs are listed in Table~\ref{complexity}, where GFLOPs represent a billion of floating-point operations per second. In addition, we analyze the time complexity of the model in terms of execution time. Among the total 114M parameters of the HPCL-Net, only 67M parameters are trainable, while the remaining 47M parameters are momentum-updated. Therefore, compared with the FFS-Net, the amount of trainable parameters of the HPCL-Net has hardly increased. Despite the increase in GFLOPs and execution time, the HPCL-Net achieves the significant improvements where $F1scores$ increases from 0.501 to 0.565 (6.4\%) and $mIoU$ increases from 0.620 to 0.651 (3.1\%), which is an acceptable trade-off.

\begin{table}[htbp]
    \centering
    \caption{Numerical results of Complexity.}
    \begin{tabular}{lcccc}
	\hline
	\multicolumn{1}{l}{Method}& Params(M) & Trainable Params(M) & GFLOPs & Time(sec/epoch)\\
	\hline
	FFS-Net & 64.55 & 64.55 & 233.91 & 67\\
	HPCL-Net & 114.09 & 67.23 & 1196.46 & 349\\
	\hline
    \end{tabular}
    \label{complexity}
\end{table}

\section{Conclusions}
In this paper, we proposed the HPCL-Net model based on feature fusion and supervised contrastive learning via HRSIs and DEM data for relic landslide detection. In the proposed HPCL-Net, HFE was designed that utilizes a dual-branch network to first extract optical and terrain features and then employs the CA mechanism to effectively fuse them after concatenating. We designed CSPC to sample hyper-pixels that contain rich semantic information as anchors and keys located in the back wall and side wall areas of landslides. In addition, a GCQC was designed, which consists of global category queues, a momentum encoder and momentum projection head, and is proven to significantly improve the performance of contrastive learning. The HPCL-Net was evaluated through extensive experiments. Compared with the reference model, HPCL-Net achieves a great improvement in detection accuracy of relic landslides with visually blurry features. The further ablation experiments, cross validation experiments and complexity analysis demonstrated the reliability of HPCL-Net with visualized improvements from heatmaps.

\section*{Declaration of competing interest}
The authors declare that they have no known competing financial interests or personal relationships that could have appeared to influence the work reported in this paper. 

\section*{Data availability}
The authors do not have permission to share data. 

\section*{Acknowledgements and Funding}
This work was supported by the National Key Research and Development Program of China [Grant No. 2021YFC3000400], and by the High Level Talent Team Project of the New Coast of Qingdao New District [Grant No. RCTD-JC-2019-06].

\appendix
\setcounter{figure}{0}
\section{Distribution Maps}
\begin{figure}[H]
    \centering
    \includegraphics[scale=0.35]{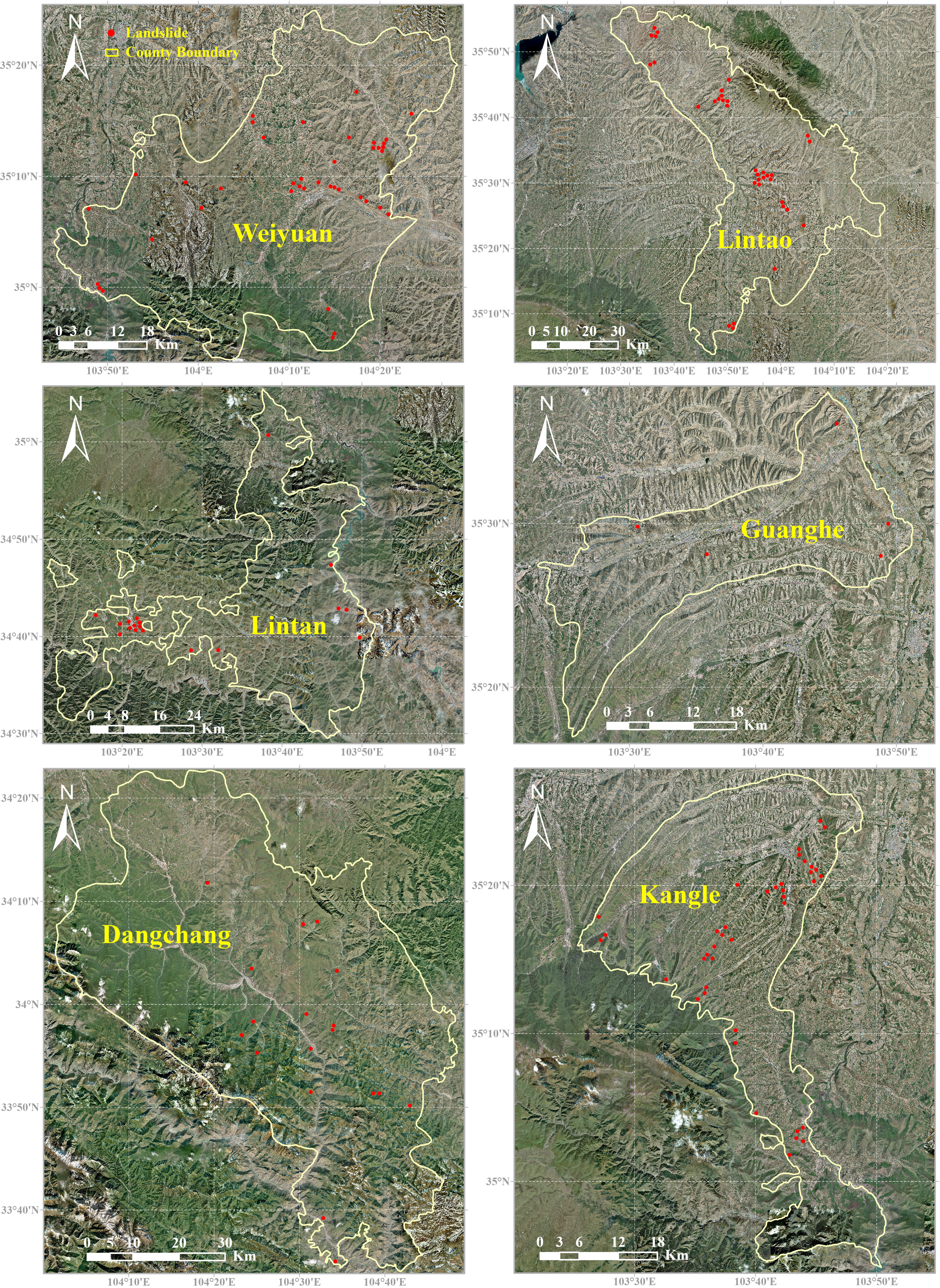}
    \DeclareGraphicsExtensions.
    \caption{The landslide distribution maps from the six counties.}
    \label{Areas}
\end{figure}

The distribution maps from Kangle, Weiyuan, Lintan, Lintao, Guanghe, and Dangchang are shown in Fig. \ref{Areas}.

\section{Training process}
The overall training process of the HPCL-Net is given in the form of PyTorch-like pseudocode.

\noindent\rule{\textwidth}{0.4pt}
\begin{verbatim}
# f1, f2: Encoder and Momentum Encoder
# p1: Projection Head
# p2: Momentum Projection Head
# m: momentum
    
# Initialize parameters
f2.params = f1.params
p2.params = p1.params

# Load a mini-batch data
for HRSI, DEM in loader:
    # Get Encoder feature map
    map1 = f1(HRSI, DEM)
    # Get Momentum Encoder feature map
    map2 = f2(HRSI, DEM)

    # Sample anchors from Encoder feature map
    anchors = anchor_selection_strategy(map1)
    # Dimensional reduction by p1
    anchors = p1(anchors)

    # Re-encode anchors to new keys
    new_keys = anchor_selection_strategy(map2)
    # Dimensional reduction by p2
    new_keys = p2(new_keys)

    # Update the queues
    enqueue(new_keys)
    dequeue(last_minibatch_keys)

    # Sample keys from the queues
    keys = key_selection_strategy(queues)
    
    # Get the categorical score map from Decoder
    score_map = decoder(map1)

    # Loss
    loss1 = SCLoss(anchors, keys)
    loss2 = CrossEntropyLoss(score_map, label)
    loss = alpha * loss1 + beta * loss2
    
    # Update Momentum Encoder and Projection Head
    f2.params = f2.params * m + f1.params * (1-m)
    p2.params = p2.params * m + p1.params * (1-m)

    # Backpropagate to update f1, p1 and Decoder
    loss.backward()
    update(f1.params, p1.params, decoder.params)
\end{verbatim}
\noindent\rule{\textwidth}{0.4pt}




\bibliographystyle{elsarticle-num-names}
\bibliography{Manuscript}
\end{document}